\documentclass[10pt,conference]{IEEEtran}
\IEEEoverridecommandlockouts
\usepackage{cite}
\usepackage{amsmath,amssymb,amsfonts}
\usepackage{graphicx}
\usepackage{textcomp}
\usepackage{xcolor}
\usepackage{orcidlink}
\usepackage{algorithmicx}
\usepackage{algorithm}
\usepackage{algpseudocode}
\usepackage{booktabs}
\usepackage{subcaption}
\usepackage{tikz}
\usepackage{pgfplots}
\usepackage{booktabs}
\usepackage{multirow}
\pgfplotsset{compat=1.17}
\usepackage{placeins}

\def\BibTeX{{\rm B\kern-.05em{\sc i\kern-.025em b}\kern-.08em
    T\kern-.1667em\lower.7ex\hbox{E}\kern-.125emX}}
\begin{document}

\title{Constraint-Driven Warm-Freeze for Efficient Transfer Learning in Photovoltaic Systems}



\author{
\IEEEauthorblockN{
Yasmeen Saeed, Ahmed Sharshar, Mohsen Guizani
}
\IEEEauthorblockA{
Mohamed bin Zayed University of Artificial Intelligence, Abu Dhabi, United Arab Emirates\\
\{Yasmeen.Saeed, Ahmed.Sharshar, Mohsen.Guizani\}@mbzuai.ac.ae\\
}
}

\maketitle

\begin{abstract}
Detecting cyberattacks in photovoltaic (PV) monitoring and MPPT control signals requires models robust to bias, drift, and transient spikes, yet lightweight enough for resource-constrained edge controllers. While deep learning outperforms traditional physics-based diagnostics and handcrafted features, standard fine-tuning is computationally prohibitive for edge devices. Furthermore, existing Parameter-Efficient Fine-Tuning (PEFT) methods typically apply uniform adaptation or rely on expensive architectural searches, lacking the flexibility to adhere to strict hardware budgets. To bridge this gap, we propose \textbf{Constraint-Driven Warm-Freeze (CDWF)}, a budget-aware adaptation framework. CDWF leverages a brief warm-start phase to quantify gradient-based block importance, subsequently solving a constrained optimization problem to dynamically allocate full trainability to high-impact blocks while efficiently adapting the remainder via Low-Rank Adaptation (LoRA). We evaluate CDWF on standard vision benchmarks (CIFAR-10/100) and a novel PV cyberattack dataset, transferring from bias pretraining to drift and spike detection. The experiments demonstrate that CDWF retains 90--99\% of full fine-tuning performance while reducing trainable parameters by up to $120\times$. These results establish CDWF as an effective, importance-guided solution for reliable transfer learning under tight edge constraints.
\end{abstract}


The code for this work is publicly available at \url{https://github.com/yasmeenfozi/Constraint-Driven-Warm-Freeze}.

\section{Introduction}

Photovoltaic (PV) systems rely on sensor-driven control and grid monitoring, making them vulnerable to signal manipulation despite exceeding 1.2 TW in global capacity with over 240 GW added in 2022 \cite{IEAPVPS2023}. In the United States, PV accounted for over 50\% of new power generation capacity in 2023 \cite{NREL2024}. This rapid growth has been accompanied by increased system interconnectivity through smart inverters and IoT-enabled monitoring, which expands the cyber-attack surface \cite{ye2022pvsecurity,Harrou2023}. Real-world incidents have already been reported, including a 2019 cyberattack that temporarily disrupted over 500~MW of renewable generation \cite{ye2022pvsecurity}. Common attack modalities include bias (persistent offsets), drift (gradual changes), and spike attacks (brief high-magnitude disturbances), which can mislead MPPT controllers and cause energy loss or grid instability.

To mitigate such threats, prior work has explored a wide range of PV cyberattack detection techniques spanning physics-based diagnostics, feature-engineered pipelines, and data-driven neural models \cite{9656555,Harrou2023,hassan2025pvcyberattacks}. Traditional approaches rely on hand-crafted time--frequency features or physical consistency checks using waveform measurements \cite{9656555}. More recent methods adopt deep learning architectures, including CNNs, LSTMs, Transformer-based hybrids, and spiking neural networks, to directly model PV monitoring signals \cite{Valderrama2025,11207726,11004263}. Unsupervised and semi-supervised approaches, such as ensemble detectors, variational autoencoders, and GAN-based models, have also been explored to identify previously unseen faults or cyber intrusions without labeled attacks \cite{9300481}. Topology-aware and hybrid optimization-based detectors further improve detection performance in grid-connected PV systems \cite{olojede2025topology,mohammed2025dual}. Despite strong progress, many PV-oriented detectors rely on specialized model designs, extensive retraining, or limited treatment of resource constraints, which complicates deployment on embedded edge hardware \cite{Paleyes_2022}.

In parallel, parameter-efficient fine-tuning (PEFT) methods reduce adaptation cost under resource constraints, with approaches such as LoRA introducing trainable low-rank matrices while freezing backbone weights \cite{hu2021loralowrankadaptationlarge}. Extensions including AdaLoRA and DoRA dynamically redistribute adaptation capacity or decompose weight updates to better approximate full fine-tuning behavior \cite{zhang2023adaloraadaptivebudgetallocation,liu2024doraweightdecomposedlowrankadaptation}. Quantization-aware approaches such as QLoRA further reduce memory requirements by combining low-rank updates with low-precision weights \cite{dettmers2023qloraefficientfinetuningquantized}. Beyond fixed adaptation rules, automated configuration search and selective parameter tuning strategies have been proposed, using Bayesian optimization, gradient importance, or sensitivity analysis to identify efficient update subsets \cite{zhou2024autopeftautomaticconfigurationsearch,lv2024parameterfinetuninglargelanguage,yu2025prunepeftiterativehybridpruning,Zhang_2024_CVPR,sahoo2025layerselectionapproachtest}. While effective, many of these approaches assume uniform adaptation across layers or rely on computationally expensive searches, limiting their practicality for edge deployment with strict and predictable hardware budgets.

\begin{figure*}[t]
    \centering
    \begin{subfigure}{0.3\textwidth}
        \centering
        \includegraphics[width=0.9\linewidth]{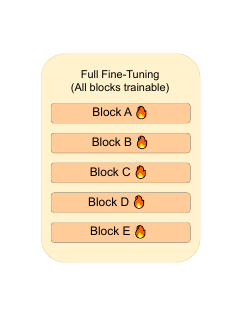}
        \caption{Full Fine-Tuning}
        \label{fig:fullft}
    \end{subfigure}
    \hfill
    \begin{subfigure}{0.33\textwidth}
        \centering
        \includegraphics[width=0.9\linewidth]{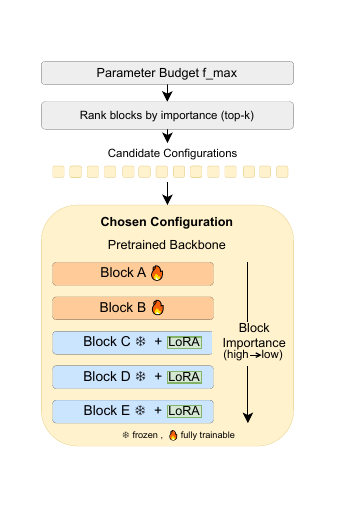}
        \caption{CDWF (Ours)}
        \label{fig:cdwf}
    \end{subfigure}
    \hfill
    \begin{subfigure}{0.3\textwidth}
        \centering
        \includegraphics[width=0.9\linewidth]{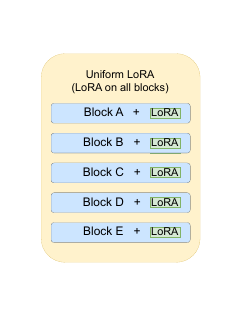}
        \caption{Uniform LoRA}
        \label{fig:lora}
    \end{subfigure}

    \caption{Comparison of adaptation strategies.
    (a) Full fine-tuning updates all blocks.
    (b) CDWF selectively adapts high-importance blocks under a parameter budget.
    (c) Uniform LoRA applies identical adapters to all blocks.}
    \label{fig:cdwf_overview}
\end{figure*}

To address these limitations, we propose Constraint-Driven Warm-Freeze (CDWF), a block-level, budget-aware fine-tuning framework for edge-based PV cyberattack detection. CDWF partitions a pretrained network into blocks and selectively determines which blocks remain fully trainable and which are frozen and adapted using lightweight LoRA modules. Block selection is guided by gradient-based importance estimates, and candidate configurations are chosen through a constrained optimization procedure that enforces a user-defined parameter budget. This allows CDWF to allocate adaptation capacity where it is most impactful while preserving efficiency elsewhere.

We evaluate CDWF on realistic PV cyberattack detection tasks involving bias, drift, and spike attacks on MPPT-controlled systems, and additionally on CIFAR-10 and CIFAR-100 image classification benchmarks to demonstrate cross-domain generality. Across all settings, CDWF is compared against full fine-tuning and strong PEFT baselines under matched parameter constraints, achieving high performance retention while training only a small fraction of parameters.

The main contributions of this work are:
\begin{itemize}
    \item A constraint-driven, block-level fine-tuning framework enabling efficient adaptation under explicit hardware budgets.
    \item A realistic PV cyberattack dataset simulating bias, drift, and spike attacks on MPPT monitoring signals.
    \item Extensive experiments across domains and backbones demonstrating improved performance over uniform PEFT baselines at comparable parameter budgets.
\end{itemize}

\section{Methodology}

CDWF enables budget-aware adaptation of pretrained models by balancing full fine-tuning and parameter-efficient adaptation. CDWF automatically identifies which layers should remain fully trainable and which can be frozen and adapted using LoRA, while satisfying a specified parameter budget. Algorithm ~\ref{alg:cdwf} outlines the procedure and Figure~\ref{fig:cdwf_overview} summarizes the CDWF workflow, showing how a parameter budget drives block ranking, candidate configuration evaluation, and the final adapted architecture.

CDWF depends on six stages: (1) a brief warm-start training to collect 
gradient information, (2) gradient analysis to identify important blocks, (3) predictive modeling of configuration performance, (4) constrained search for an optimal architecture, (5) configuration application and lastly (6) fine-tuning.

\subsection{Problem Formulation}

The main challenge is determining which layers need full trainability and which can use efficient LoRA adaptation. This decision revolves around specified parameter constraint and the task's requirements. Rather than selecting this manually, we formulate the decision as an optimization problem solved automatically.

We represent each adaptation strategy as a configuration $\mathcal{C} = (\mathcal{K}, \mathcal{F}, r)$. Where $\mathcal{K}$ is the set of blocks kept fully trainable, $\mathcal{F}$ is the set of blocks frozen with LoRA adaptation, and $r$ is the LoRA rank applied.

Given $B$ total blocks, we construct the candidate set by varying the number of kept 
blocks $k \in \{0, 1, \ldots, B\}$ and selecting the top-$k$ most important 
blocks based on gradient magnitude (computed in Section~\ref{sec:importance}). 
The remaining $B-k$ blocks are frozen and adapted using LoRA.

This construction results in a linear search space over the number of trainable blocks, avoiding the exponential $2^B$ subset space. By restricting the selection to the $\textit{top-k}$ most important blocks, the approach remains computationally efficient while ensuring that adaptation capacity is allocated to the most task-relevant parts of the model.

Given a maximum parameter budget  $f_{\text{max}}$. We want the most accurate configuration $\mathcal{C}^*$ that respects the set limit:
\begin{equation}
\mathcal{C}^* = \arg\max_{\mathcal{C}} \hat{A}(\mathcal{C}) 
\quad \text{subject to} \quad 
f_{\text{train}}(\mathcal{C}) \leq f_{\text{max}}
\label{eq:optimization}
\end{equation}

where $\hat{A}(\mathcal{C})$ is the predicted accuracy and $f_{\text{train}}(\mathcal{C})$ is the fraction of trainable parameters.

Evaluating all candidate configurations through full training would be computationally expensive. The following sections describe how this is addressed using gradient-based importance estimation and predictive modeling.

\subsection{Brief Warm-Start}

To bootstrap the adaptation process and collect gradient information for importance estimation, we first perform a brief warm-start phase with minimal additional overhead where all model parameters are trainable. This initial adaptation enables the identification of blocks that require greater updates for the target task, rather than measuring gradients on a model still specialized for its original pretraining objective.

We perform a brief warm-start training phase for $E_{warm}$ epochs, serving three purposes: (1) it establishes a baseline performance of validation accuracy $A_{\text{warm}}$ that represents the first task's adaptation, (2) it generates gradient signals during backpropagation indicating which blocks require more parameter updates for the targeted task, and (3) it provides an improved initialization that speeds up convergence during fine-tuning.

\subsection{Gradient-Based Block Importance Estimation}
\label{sec:importance} 
Following the warm-start phase, we determine which blocks should remain fully trainable and which should be adapted using LoRA. The factor indicating this is the gradient magnitudes where blocks requiring substantial weight changes to minimize loss will have larger gradients, in contrast blocks that are well-suited will show smaller gradients.

We measure gradient activity on validation data during warm-start (performed only during training), tracking how much each block's parameters need to change. For each block $i$, we collect the gradient norms \cite{chen2018gradnormgradientnormalizationadaptive} across $N$ validation batches:

\begin{equation}
G_i = \frac{1}{N} \sum_{j=1}^{N} \left\|\nabla_{\theta_i} \mathcal{L}(f_{\theta^{\text{warm}}}(x_j), y_j)\right\|_2
\label{eq:gradient_norm}
\end{equation}

where $\theta_i$ are the parameters of block i, $f_{\theta^{\text{warm}}}$ is the warm-started model, $\mathcal{L}$ is the cross entropy loss, and $(x_j, y_j)$ are validation input-label pairs. This provides a magnitude score for each block (higher values = more updating needed). We compute these scores on validation data to reflect generalization rather than memorization, and normalize them to sum to one for direct comparison:

\begin{equation}
I_i = \frac{G_i}{\sum_{k=1}^{B} G_k}
\label{eq:importance}
\end{equation}

where $I_i \in [0, 1]$ represents block $i$'s importance. Importance scores guide which blocks to keep trainable versus freeze with LoRA. CDWF selects configurations that maximize accuracy while satisfying the user specified constraints. The final architecture choice will balance these importance scores with the constraints.

\begin{algorithm}[t]
\caption{Constraint-Driven Warm-Freeze (CDWF)}
\label{alg:cdwf}
\begin{algorithmic}[1]
\Require Pretrained model $\theta_{\text{pretrained}}$, budget $f_{\max}$, reference accuracy $A_{\text{ref}}$, warm-start epochs $E_{\text{warm}}$, fine-tuning epochs $E_{\text{ft}}$
\Ensure Fine-tuned model $\theta^*$

\State $(\theta_{\text{warm}},A_{\text{warm}})\gets \textsc{WarmStart}(\theta_{\text{pretrained}},E_{\text{warm}})$
\State $\{I_i\}_{i=1}^{B}\gets \textsc{ComputeImportance}(\theta_{\text{warm}})$
\State $G_{\max}\gets A_{\text{ref}}-A_{\text{warm}},\;\mathcal{V}\gets\emptyset$
\For{each configuration $C=(K,F,r)$ in candidates}
    \State $\hat{A}(C)\gets A_{\text{warm}} + G_{\max}\!\left(\sum_{i\in K} I_i + \sum_{j\in F}\eta(r)I_j\right)$
    \State $f_{\text{train}}(C)\gets P_{\text{train}}(C)/P_{\text{total}}$
    \If{$f_{\text{train}}(C)\le f_{\max}$}
        \State $\mathcal{V}\gets \mathcal{V}\cup\{(C,\hat{A}(C))\}$
    \EndIf
\EndFor
\State $C^*\gets \arg\max_{(C,\hat{A})\in\mathcal{V}} \hat{A}$
\State \Return $\textsc{FineTune}(\theta_{\text{warm}},C^*,E_{\text{ft}})$
\end{algorithmic}
\end{algorithm}

\subsection{Predictive Accuracy Model}

Given the estimated importance scores, we aim to predict the performance of each configuration without explicitly training it, as exhaustively training all possible configurations would be computationally expensive.
Instead, a simple model is used to estimate accuracy.

Warm-start provides us with baseline accuracy $A_{\text{warm}}$. CDWF avoids the costly training of all configurations and instead calibrates the predictor using a single reference run, which in experiments is Full Fine-Tuning (Full-FT), following established fine-tuning practices \cite{howard2018universallanguagemodelfinetuning}.

We estimate the reference improvement relative to warm-start as
$G_{\max} = A_{\text{ref}} - A_{\text{warm}}$,
where $A_{\text{ref}}$ is the accuracy from a single reference run (Full-FT) used only for
calibration and for reporting retention. For each model–dataset pair, we obtain $A_{\text{ref}}$ from a single offline Full-FT run performed once per model–dataset pair for calibration, and it is not part of the constrained adaptation process. A mixed configuration is expected to achieve a fraction of this gain depending on how capacity is allocated between fully trainable blocks and LoRA-adapted blocks.

LoRA provides partial adaptation in comparison to full trainability, and we model its 
effectiveness with an efficiency factor $\eta(r) = \min(0.5, r/8)$, accounting for 
both the diminishing returns observed beyond rank 4~\cite{hu2021loralowrankadaptationlarge} 
and the task-dependent nature of LoRA.

For any configuration $\mathcal{C} = (\mathcal{K}, \mathcal{F}, r)$, we predict final accuracy as:

\begin{equation}
\hat{A}(C) = A_{\text{warm}} + G_{\max} \cdot 
\left(\sum_{i \in \mathcal{K}} I_i + \sum_{j \in \mathcal{F}} \eta(r) \cdot I_j\right)
 \label{eq:predicted_acc}
\end{equation}
where the term $\sum_{i \in \mathcal{K}} I_i$ is the total importance contribution from fully trainable blocks, and the second term $\sum_{j \in \mathcal{F}} \eta(r) \cdot I_j$ represents the weighted 
importance contribution from the LoRA applied blocks.

\subsection{Parameter-Aware Constrained Optimization}

Using the predicted accuracy model, CDWF selects an adaptation configuration that satisfies the user-specified parameter budget. For each candidate configuration $C=(\mathcal{K},\mathcal{F}, r)$, we compute the fraction of trainable parameters
$f_{train}(C)=\frac{P_{train(C)}}{P_{total}}$
,where trainable parameters include the classification head, all parameters in the fully trainable blocks $\mathcal{K}$, and the LoRA parameters introduced in the frozen blocks $\mathcal{F}$. 

Configurations with $f_{train}(C) > f_{max}$ are discarded. Among the remaining feasible configurations, CDWF selects the one with the highest predicted accuracy using equation ~\ref{eq:optimization}. 

Importantly, the parameter budget acts as an upper bound rather than a target. If increasing the number of trainable blocks does not yield a sufficient predicted accuracy gain, CDWF retains a smaller configuration even when additional budget is available. As a result, multiple budget values may map to the same architecture until the predicted benefit of increasing trainable depth justifies the use of additional parameters. Once the best configuration is found, it is applied it to the model.

\textbf{Applying the Configuration:} We modify the warm-started model 
based on the chosen configuration: blocks in $\mathcal{K}^*$ stay 
trainable, while blocks in $\mathcal{F}^*$ get frozen and fitted with 
LoRA layers at rank $r^*$. For frozen blocks, we keep their original 
weights locked and add small LoRA adaptation matrices on top. The 
classification head always stays trainable.

\textbf{LoRA Implementation:} 
For the blocks selected to be frozen, we apply LoRA only to the main spatial operation in each block. In CNNs, this corresponds to the second convolution (conv2), which is the $3\times3$ convolution in ResNet-50 bottleneck blocks or the $3\times1$ convolution in BasicBlock, and is responsible for learning spatial features. In Vision Transformers, we apply LoRA to the self attention projections.

\section{Dataset}

\begin{table*}[t]
\centering
\caption{Performance comparison of CDWF under a forced rank constraint:}
\label{tab:main_results}
\begin{tabular}{l|l|cccccc}
\toprule
\textbf{Dataset} & \textbf{Method} & 
\textbf{Test Acc(\%)} & \textbf{Test AUC(\%)} & \textbf{Retention(\%)} & \textbf{Param Count} & \textbf{Param Reduction} \\
\midrule
CIFAR-10 & Full-FT & 88.28 & -- & 100.0 & 23,528,522 & 1.0$\times$ \\
         & LoRA (rank=8) & 81.29  & -- & 92.08 &  457,738 & 52.36$\times$ \\
         & CDWF (rank=8) & \textbf{85.29} & -- & \textbf{96.61} & \textbf{401,994}  & \textbf{58.5$\times$} \\
\midrule
CIFAR-100 & Full-FT & 64.99 & -- & 100.0 & 23,528,522 & 1.0$\times$ \\
         & LoRA (rank=16) &  49.32  & -- & 75.89 & \textbf{809,060}  & \textbf{29.1$\times$} \\
         & CDWF (rank=16) & \textbf{59.57} & -- & \textbf{91.66} &  818,596 & 28.7$\times$  \\
\midrule
Drift & Full-FT & 91.88 & 98.49 & 100.0 & 3,844,930 & 1.0$\times$ \\
         & LoRA (rank=16) &  86.48 & 95.69 & 97.15 & \textbf{123,906}  & \textbf{31.0$\times$}\\
         & CDWF (rank=16) & \textbf{90.10} & \textbf{97.71} & \textbf{99.20} &  124,482 & 30.9$\times$  \\
\midrule
Spike & Full-FT & 94.72 & 99.29 & 100.0 & 3,844,930 & 1.0$\times$ \\
         & LoRA (rank=4) & 93.58  & 98.86 & 99.56 & \textbf{31,746}  & \textbf{121$\times$} \\
         & CDWF (rank=4) & \textbf{94.10} & \textbf{99.17} & \textbf{99.88} &  32,322 & 119$\times$ \\
\bottomrule
\end{tabular}
\addvspace{4pt}
\textit{Note:} The LoRA rank is fixed to the best-performing LoRA baseline for each dataset in order to closely match trainable parameter counts. 
Adaptive CDWF behavior is analyzed separately in Tables II and IV.
\end{table*}

To evaluate CDWF in resource-constrained PV edge settings, we generate three
time-series cyberattack datasets for PV MPPT monitoring:
\textbf{Bias} attacks (used for pre-training), and two target adaptation tasks:
\textbf{Drift} attacks and \textbf{Spike} attacks. 
Each dataset is a balanced binary classification problem over 10-second voltage
snippets, containing both normal and attacked versions of the same
underlying operating condition.

\textbf{PV simulation and normal traces:}
We build a PV simulator based on single-diode model principles \cite{Elhammoudy2024} and generate
diverse operating conditions by varying solar geometry, weather regimes,
temperature cycles, and electrical parameters\cite{10224249}.
We generate $14{,}400$ normal snippets, each of length 10 seconds at 30 Hz
($300$ samples per snippet), resulting in 40 hours of normal operation.

\textbf{Attack framework shared across all attacks:}

\begin{figure}[t]
    \centering
    \begin{subfigure}{0.48\textwidth}
        \centering
        \includegraphics[width=\linewidth]{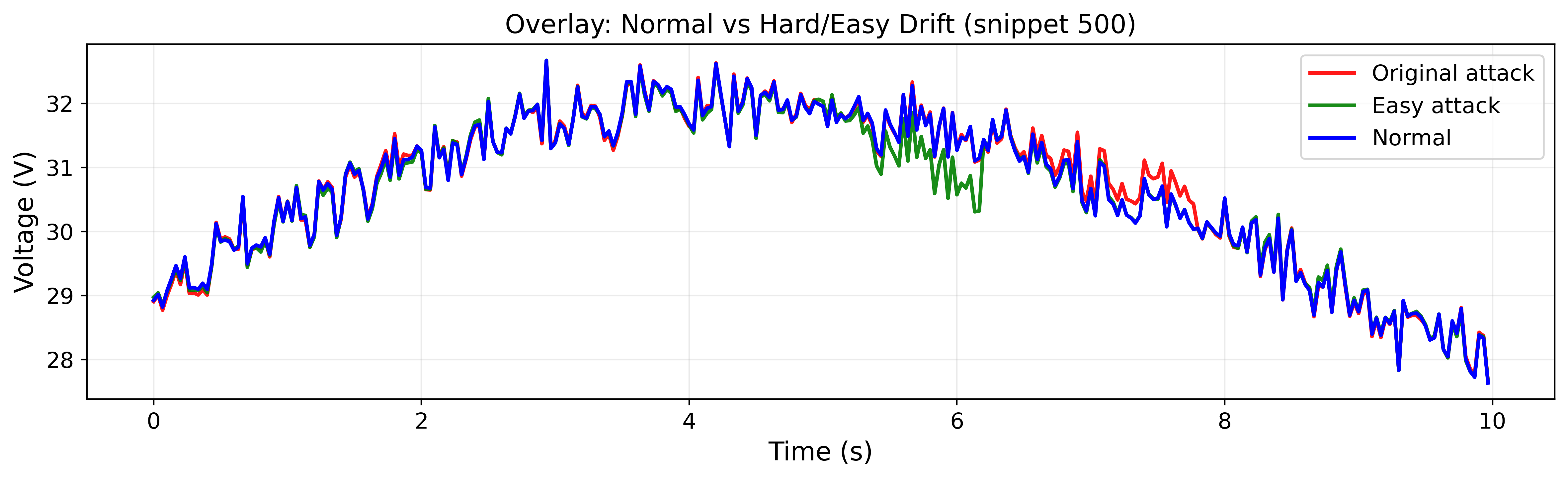}
        \caption{Drift attack}
    \end{subfigure}\hfill
    \begin{subfigure}{0.48\textwidth}
        \centering
        \includegraphics[width=\linewidth]{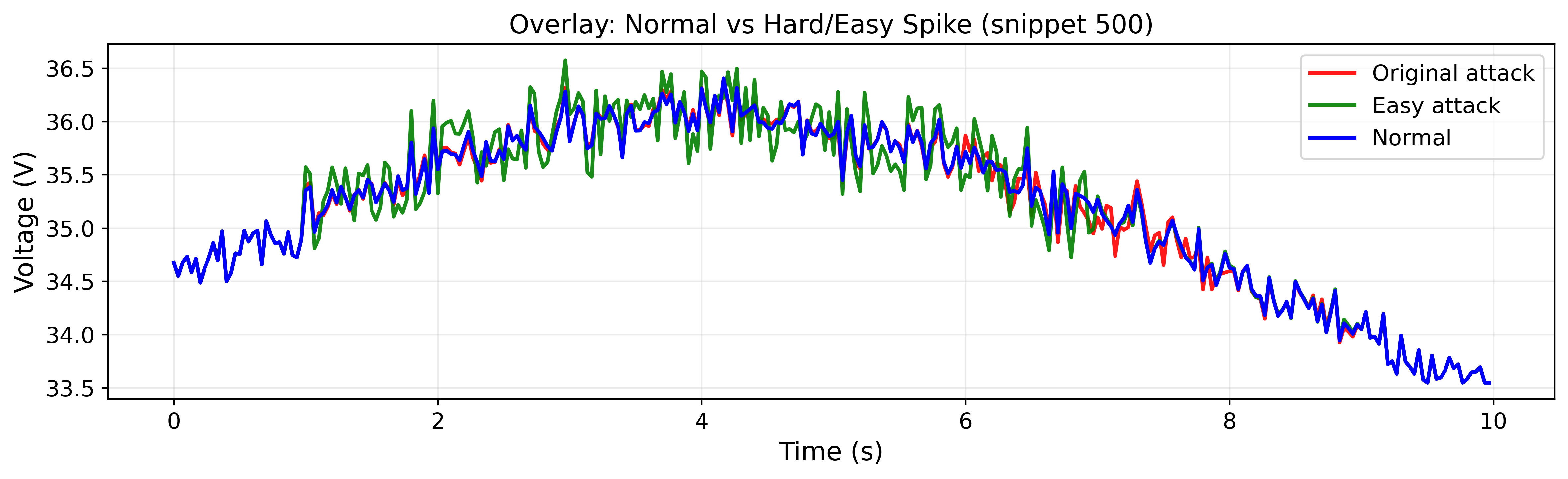}
        \caption{Spike attack}
    \end{subfigure}
    \caption{10-second voltage traces showing (a) drift attack 
with easy and harder gradual voltage deviation and (b) spike attack with easy and harder transient 
disturbances.}
    \label{fig:snippets}
\end{figure}

All attacks are generated to closely mimic realistic disturbances rather than fake anomalies. Specifically, the first and last second of each signal are left untouched to preserve the natural boundary, while the attack is limited to a randomly selected interior window with a minimum duration. The start/end time and duration of attacks vary across samples, preventing fixed or easily detectable patterns. Figure~\ref{fig:snippets} compares the original attacks used in this work with an easier variant shown only for illustration. The easier attacks exhibit larger and more obvious deviations, whereas the hard attacks are subtler and differ in magnitude, timing, and duration, making them harder to distinguish from normal behavior.

\textbf{Bias attacks:}
Bias attacks simulate subtle measurement manipulation through small
multiplicative perturbations applied inside $[X_s, Y_s)$.
In the proposed implementation, each attack sample in the interval is scaled by a
random signed factor in the range $\pm(0.3\%\text{--}0.8\%)$ (with mild additive
noise), while the edges remain identical to the original snippet.
This creates a challenging case where the mean voltage distributions of normal
and attacked samples heavily overlap, forcing the detector to use temporal
patterns rather than trivial thresholds.

\textbf{Drift attacks:}
Drift attacks simulate gradual sensor degradation or stealthy integrity attacks
by applying a linear multiplicative ramp within $[X_s, Y_s)$.
The ramp starts near 0 and increases smoothly to a final magnitude sampled in
$0.5\%\text{--}1.5\%$, with a random direction. This produces slow, structured
changes that are intentionally hard to separate from natural operating variation.

\textbf{Spike attacks:}
Spike attacks simulate brief high-frequency disturbances (e.g., electromagnetic interference or fast injection events) by injecting sparse signed impulse perturbations within $[X_s, Y_s)$. In each snippet, 3–10 spikes are introduced, each spanning up to 4 samples, with impulse magnitudes sampled uniformly from 1\%–20\% of the nominal signal level (normalized voltage units) and superimposed on mild background noise $(\sigma=0.002)$. This produces short irregularities without inducing a sustained shift in the overall signal distribution.

\textbf{Paired attack generation:}
For every normal snippet ID $i$, we generate a paired attacked snippet using the
same underlying trace, producing $14{,}400$ attacked snippets and
$28{,}800$ total samples (80 hours total).
To avoid leakage, each snippet ID is assigned to exactly one split, and both its
normal and attacked versions are kept in the same split.

\textbf{Train/val/test split:}
We split the $14{,}400$ snippet IDs into 70\% training (10,080 pairs), 15\%
validation (2,160 pairs), and 15\% testing (2,160 pairs). This ensures the model does not memorize an operating condition from the normal version and recognize it during testing via the attacked counterpart.

 Attacks are kept low in magnitude and applied only within interior time intervals, so detection cannot rely on obvious boundary effects or simple mean shifts. This setup reflects realistic photovoltaic behavior, where measurements can naturally show bias-like offsets, gradual drifts, and short spikes due to environmental changes, control actions, and system disturbances \cite{ye2022pvsecurity}.

\section{Experimental Setup}

\begin{table*}[t]
\centering
\caption{Budgeted performance of CDWF across datasets under explicit trainable-parameter budgets $f_{\max}\in\{0.02,0.05,0.10\}$. Retention is computed relative to Full Fine-Tuning on the same dataset (for PV tasks, retention is computed using AUC).}
\label{tab:budget_sweep}
\small
\setlength{\tabcolsep}{6pt}
\begin{tabular}{llccccc}
\toprule
Dataset & $f_{\max}$ &
Test Acc (\%) & Test AUC (\%) &
Retention (\%) &
Trainable (\%) &
Chosen $(k,r)$ \\
\midrule

\multirow{3}{*}{CIFAR-10} 
& 0.02 & 85.47 & --    & 96.82 & 1.07  & (1,4) \\
& 0.05 & 86.13 & --    & 97.56 & 2.65  & (2,4) \\
& 0.10 & 87.14 & --    & 98.71 & 9.0  & (3,4) \\
\midrule

\multirow{3}{*}{CIFAR-100} 
& 0.02 & 59.53 & --    & 91.60 & 1.84  & (1,4) \\
& 0.05 & 59.53 & --    & 91.60 & 1.84  & (1,4) \\
& 0.10 & 61.82 & --    & 95.12 & 9.71  & (3,4) \\
\midrule

\multirow{3}{*}{PV Drift} 
& 0.02 & 90.33   & 97.93  & 99.41 & 1.70 & (2,2) \\
& 0.05 & 90.45   & 98.00  & 99.49 & 4.15 & (3,4) \\
& 0.10 & 90.57   & 98.05  & 99.55 & 6.65 &  (4,4) \\
\midrule

\multirow{3}{*}{PV Spike} 
& 0.02 & 94.24   & 99.16 & 99.87 & 1.45 & (1,4) \\
& 0.05 & 94.48   & 99.21 & 99.92 & 4.15 & (3,4) \\
& 0.10 & 94.69   & 99.24 & 99.95 & 6.65 & (4,4) \\
\bottomrule
\end{tabular}
\addvspace{4pt}
\textit{Note: (k,r) = (k is the number of trainable blocks, r is the LoRA rank on frozen blocks)}
\end{table*}

\subsection{Setup}
We evaluated CDWF on multiple datasets and models to showcase its generalization in different domains and model architectures: CIFAR-10~\cite{Krizhevsky09learningmultiple}, CIFAR-100~\cite{Krizhevsky09learningmultiple} pretrained on ImageNet, and time-series cyber-attack detection 
for PV MPPT systems-Drift and Spike pretrained on Bias.

\textbf{Benchmark Datasets:} To demonstrate the 
generalizability of CDWF across different architectures and tasks, we evaluate on standard image classification benchmarks: CIFAR-10 and CIFAR-100.
For CIFAR datasets, we apply the standard split from the original training set.

\textbf{Training Configuration:} All experiments were conducted using a fixed random seed (42).
We train Full Fine-Tuning for 10 epochs using AdamW optimizer~\cite{loshchilov2019decoupled} with weight decay $\lambda = 10^{-2}$ and a maximum learning rate of
$3 \times 10^{-4}$. The batch size is 64 for all experiments and were run on a single NVIDIA RTX 5000 Ada Generation GPU.

\textbf{Baselines and Experimental Setup:} Unless otherwise mentioned, we use ResNet architectures in all experiments. Specifically, ResNet-50 for CIFAR experiments, which consists of 16 Bottleneck blocks and is pretrained on ImageNet, and a 1D ResNet for drift and spike detection, composed of 8 BasicBlock layers and pretrained on bias attack detection. Blocks are selected in a ranked, cumulative manner based on importance, where tighter budgets retain only the most important blocks and larger budgets progressively include additional blocks. We compare CDWF against two baselines: full fine-tuning, where all parameters are trainable and which also serves as the reference run for computing retention, and LoRA-only baselines evaluated at ranks \{1, 2, 4, 8, 16\}.

\textbf{Constraints:} We tested on three parameter budgets: $f_{\text{max}} \in \{0.02, 0.05, 0.10\}$ (2\%, 5\%, 10\% trainable parameters). All experiments use adaptive LoRA ranks $r \in \{1, 2, 4,8, 16\}$.

\textbf{Evaluation Metrics:} Metrics reported are test accuracy, retention (relative to the reference run), parameter reduction, and the chosen architecture. For Drift and Spike detection, we report test accuracy in the table, but retention is computed using test AUC as it is the primary metric for binary classification tasks.

\begin{table*}[t]
\centering
\caption{Effect of LoRA rank on CIFAR-10 under a fixed 10-epoch budget 
(3 warm-start + 7 fine-tuning epochs). CDWF is evaluated with fixed LoRA ranks 
and with adaptive rank selection.}
\label{tab:cifar10_rank_ablation}
\begin{tabular}{l|c c c c c c }
\toprule
\textbf{Method} & \textbf{Rank} & 
\textbf{Test Acc (\%)} & \textbf{Retention (\%)} &
\textbf{Trainable (\%)}  & 
\textbf{Pred. Error (\%)} & \textbf{Architecture} \\
\midrule
Full-FT & -- & 88.28 & 100.0 & 100.0  & -- & All trainable \\

\midrule
LoRA-10 & 1  & 76.11 & 86.21 & 0.32 & -- & All LoRA-r1 \\
LoRA-10 & 2  & 78.34 & 88.74 & 0.55  & -- & All LoRA-r2 \\
LoRA-10 & 4  & 79.89 & 90.50 & 1.01  & -- & All LoRA-r4 \\
LoRA-10 & 8  & 81.29 & 92.08 & 1.91  & -- & All LoRA-r8 \\
LoRA-10 & 16 & 81.23 & 92.01 & 3.67  & -- & All LoRA-r16 \\

\midrule
CDWF-10 & 1  & 85.61 & 96.98 & 0.60  & 0.40 & 1K + 15L-r1  \\
CDWF-10 & 2  & 85.22 & 96.53 & 0.76 & 0.37 & 1K + 15L-r2 \\
CDWF-10 & 4  & 85.47 & 96.82 & 1.07 & 0.88 & 1K + 15L-r4 \\
CDWF-10 & 8  & 85.29 & 96.61 & 1.69 & 1.06 & 1K + 15L-r8 \\
CDWF-10$^\dagger$ & 16 & 85.74 & 97.12 & 2.90 & 0.61 & 1K + 15L-r16 \\

\midrule
CDWF-Adaptive & $\{1,2,4,8,16\}$ & 85.47 & 96.82 & 1.07 & 0.88 & 1K + 15L-r4 \\
\bottomrule
\end{tabular}

\vspace{4pt}
\footnotesize
\textit{Note:} 
CDWF uses a parameter constraint of $f_{\max}=0.02$ for all forced-rank runs, except $^\dagger$CDWF-r16 where $f_{\max}=0.03$ was required for feasibility. 
Architecture notation: $X$K = $X$ blocks kept fully trainable; $Y$L-r$Z$ = $Y$ blocks frozen with LoRA rank $Z$.
\end{table*}






\section{Results and Analysis}

\subsection{Main Results}

Table~\ref{tab:main_results} compares CDWF with Full Fine-Tuning and uniform LoRA across all four datasets under closely matched parameter budgets. For each dataset, we first identify the strongest LoRA baseline and then constrain CDWF to use a similar number of trainable parameters, ensuring that improvements cannot be attributed to additional capacity. Under this controlled comparison, CDWF mostly achieves higher accuracy and retention than LoRA at the same rank, while using a similar parameter count. A key reason for this behavior is that CDWF selectively allocates full trainability to high-importance blocks, while the remaining blocks are adapted using low-rank updates, rather than uniformly distributing adaptation capacity across all blocks.

This trend is most visible on more challenging datasets and photovoltaic tasks. On CIFAR-100, CDWF substantially improves over uniform LoRA even when no blocks are fully unfrozen, indicating that the gains are not only due to block unfreezing but also due to warm-start initialization and the placement of LoRA. On the PV drift and spike detection tasks, CDWF remains close to the full fine-tuning operating point and consistently improves upon LoRA in both accuracy and AUC. These results show that selectively placing adaptation capacity, rather than spreading it uniformly, leads to more reliable performance across datasets.

Table~\ref{tab:budget_sweep} shows that CDWF adjusts its capacity in response to the available parameter budget, retaining only the most important blocks under tighter constraints and progressively adding blocks as the budget increases. For example, on CIFAR-10, increasing the budget from 0.02 to 0.1 improves test accuracy from 85.47\% to 87.14\%, with CDWF correspondingly selecting deeper trainable configurations only when enough capacity is available.

Across the remaining tasks, a similar pattern is shown. CIFAR-100 reveal a flat response at lower budgets followed by a clear improvement once additional blocks can be meaningfully adapted, reflecting the nature of blocks chosen. On the photovoltaic drift and spike tasks, CDWF remains close to the full fine-tuning operating point even at small budgets and shows gradual gains as the budget increases, indicating stable and predictable adaptation behavior.

\begin{figure}[t]
    \centering
    \includegraphics[width=\linewidth]{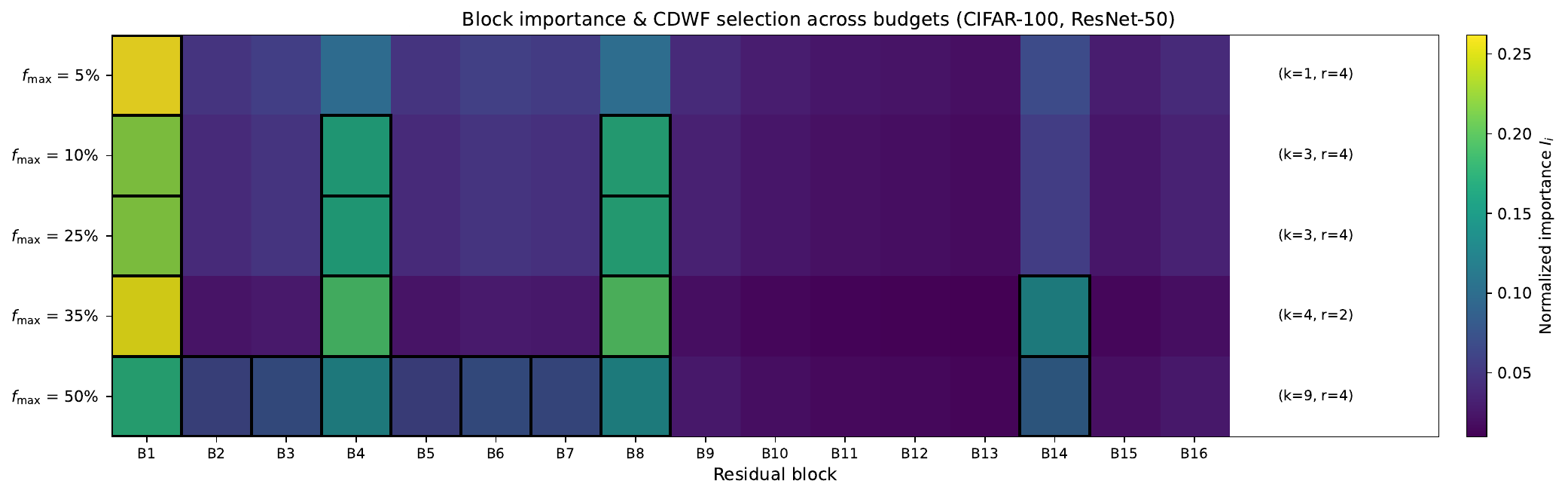}
    \caption{Block importance and CDWF block selection across budgets on CIFAR-100. }
    \label{fig:heatmap}
\end{figure}

\subsection{Ablation}

Table~\ref{tab:cifar10_rank_ablation} examines the effect of LoRA rank under a fixed 10-epoch training budget. For uniform LoRA, increasing the rank steadily improves accuracy, but performance saturates well below full fine-tuning. In contrast, CDWF exhibits weak sensitivity to rank choice across the same range, with performance remaining nearly identical across ranks, and CDWF configurations generally outperform their LoRA counterparts at the same rank. This behavior follows directly from how CDWF allocates adaptation capacity. Rather than relying on higher LoRA ranks to recover accuracy, CDWF benefits primarily from warm-start initialization and importance-guided block selection. Figure~\ref{fig:heatmap} illustrates this mechanism on CIFAR-100, where CDWF consistently concentrates capacity on a small subset of high-importance blocks while leaving less relevant blocks frozen and adapted with LoRA. As a result, increasing LoRA rank alone yields diminishing returns, leading to stable performance across the range of ranks.

The CDWF-Adaptive configuration selects the LoRA rank by maximizing the predicted accuracy under the parameter constraint, prior to fine-tuning. Although a lower rank may occasionally achieve a slightly higher test accuracy after training, such differences are within normal run-to-run variability. The adaptive choice reflects a stable operating point that provides strong predicted performance while avoiding feasibility or constraint violations observed at higher ranks.

A key component of the CDWF is how the total training budget is split between warm-start and fine-tuning, rather than the total number of epochs alone. Table~\ref{tab:ws_sensitivity} showcases this effect on CIFAR-100 by fixing the total training budget and parameter budget, while varying only the number of warm-start epochs. Under the same selected architecture, reallocating epochs from fine-tuning to warm-start leads to a substantial improvement in performance, indicating that early full adaptation plays a critical role in identifying and initializing task relevant features.

\begin{table}[t]
\centering
\caption{Warm-start sensitivity of CDWF on CIFAR-100 under a fixed parameter budget ($f_{\max}=0.05$) and a fixed total training budget of 10 epochs.}
\label{tab:ws_sensitivity}
\begin{tabular}{c c c c}
\toprule
Warm-start (epochs) & Test Acc. (\%) & Retention (\%) & Time (s) \\
\midrule
1 & 51.97 & 79.97 & 103.50 \\
2 & 56.62 & 87.12 &  103.77\\
3 & 59.60 & 91.71 & 105.27 \\
4 & 61.64 & 94.84 & 109.34 \\
\bottomrule
\end{tabular}
\end{table}

\begin{figure}[t]
    \centering
    \includegraphics[width=0.95\linewidth]{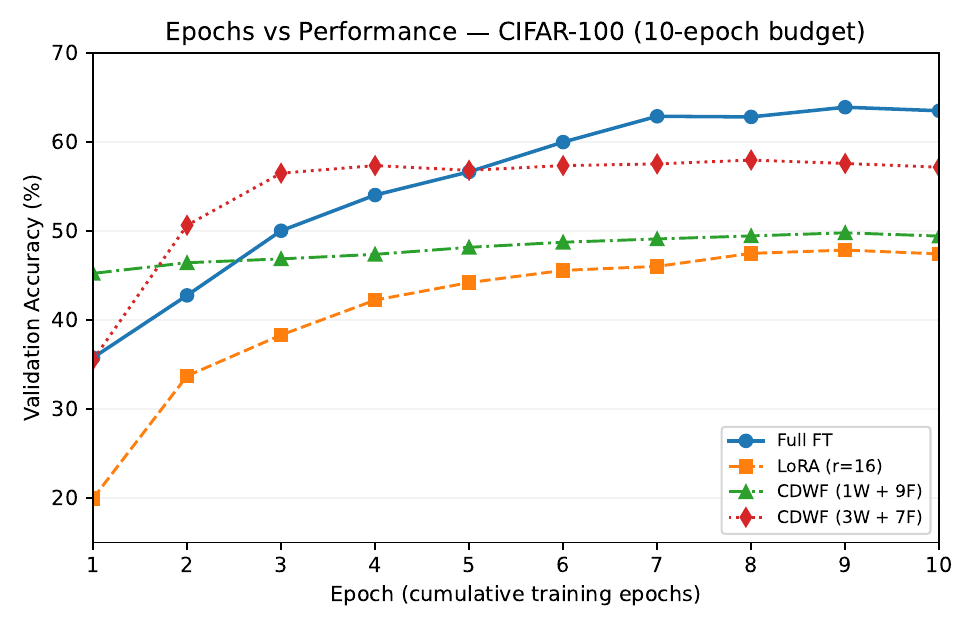}
    \caption{Epoch-wise CIFAR-100 validation accuracy (10 epoch budget). kW+(10$-$k)F indicates k warm-start epochs followed by fine-tuning epochs}
    \label{fig:chart}
\end{figure}

This effect is also visible in Figure ~\ref{fig:chart}. While full fine-tuning continues to improve steadily over the entire training budget and LoRA improves slowly before saturating, the CDWF reaches a strong operating point much earlier. Configurations with longer warm-start phases improve rapidly during the initial epochs and then stabilize during fine-tuning, achieving a better performance with the same budget. 

As demonstrated in Table~\ref{tab:backbones}, the CDWF is not tied to a specific architecture or dataset. The same behavior observed on ResNet-50 extends to smaller and larger CNN backbones as well as ViT architectures. Across all settings, the CDWF achieves near full fine-tuning performance while reducing parameters and outperforming LoRA on CNN backbones and PV tasks, yet remains competitive on ViT architectures. These results indicate that the CDWF generalizes across backbones without requiring any specific tuning.

Finally, Table~\ref{tab:vit_peft_cifar100} compares standard PEFT methods (LoRA, DoRA, and AdaLoRA) on ViT-B/16 for CIFAR-100 under the same 10-epoch budget, evaluated relative to the full fine-tuning. While all PEFT methods slightly exceed full fine-tuning accuracy while updating only a small fraction of parameters, LoRA achieves the lowest training time, with the lowest parameter count.

Across all ablations, CDWF’s improvements come from how the training budget is allocated, not from increasing the rank or trainable parameters. Warm-start initialization and importance-guided block selection consistently lead to stable gains across datasets and architectures.

\begin{table}[t]
\centering
\caption{Comparison of Accuracy and Parameter Reduction across different backbones (ResNet-18, ResNet-101, ViT). 'FL' denotes Full Fine-tuning on CIFAR-10}
\label{tab:backbones}
\resizebox{\linewidth}{!}{%
\begin{tabular}{lcccccc}
\toprule
\multirow{2}{*}{Method} & \multicolumn{2}{c}{ResNet-18} & \multicolumn{2}{c}{ResNet-101} & \multicolumn{2}{c}{ViT-B/16} \\
\cmidrule(lr){2-3} \cmidrule(lr){4-5} \cmidrule(lr){6-7}
 & Acc (\%) & Params Red. & Acc (\%) & Params Red. & Acc (\%) & Params Red. \\
\midrule
FL (Full Fine-tuning)   & 85.63 & 1.0× & 86.94 & 1.0× & 96.33 & 1.00× \\
LoRA                    & 77.43 & 21.14× & 81.74 & 65.12× & 97.92 & 860.28× \\
CDWF ($f_{\max}$ =0.05)  & 83.19 & 24.34× & 84.79 & 53.03× & 96.81 & 859.35× \\
\bottomrule
\end{tabular}%
}
\end{table}

\section{Limitations}
The CDWF uses a single offline Full Fine-Tuning run per model–dataset pair solely for predictor calibration and not as part of constrained adaptation. Block importance is estimated using a brief warm-start phase and a fixed number of validation batches, without an exhaustive sensitivity analysis over alternative importance metrics or batch counts. The LoRA efficiency factor $\eta(r)$ is a simple heuristic used only to rank candidate configurations and does not reflect the final model accuracy. Future work will aim to remove these dependencies through reference-free and adaptive strategies.

\section{Conclusion}
We presented a Constraint-Driven Warm-Freeze (CDWF), a block-level transfer learning framework that adapts pretrained models under an explicit parameter budget. By combining a short warm-start phase, gradient-based block importance estimation, and a lightweight predictive model, CDWF automatically selects which blocks remain fully trainable and which are adapted via low-rank updates, without exhaustively evaluating candidate configurations. Across image classification and time-series PV detection tasks, the proposed CDWF achieves a strong performance retention while training only a small fraction of model parameters. Compared to uniform LoRA baselines at comparable budgets, the CDWF attains a higher accuracy by allocating capacity where it is most needed, demonstrating that block-level selection is a practical alternative to rank tuning for PEFT.

A key takeaway is that performance depends not only on the size of the training budget but also on how it is allocated: warm-start signals guide the CDWF to allocate trainable capacity to influential blocks and avoid unnecessary parameter growth. Future work includes extending the CDWF beyond fixed training schedules by automatically allocating epochs based on configuration and efficiency targets, as well as validating on real edge hardware and in heterogeneous or distributed learning settings.

\begin{table}[t]
\centering
\caption{Comparison of parameter-efficient fine-tuning methods on CIFAR-100 using ViT-B/16. 
Retention is measured relative to full fine-tuning.}
\label{tab:vit_peft_cifar100}
\resizebox{\linewidth}{!}{
\begin{tabular}{lcccccc}
\toprule
\textbf{Method} 
& \textbf{Test Acc (\%)} 
& \textbf{Retention (\%)} 
& \textbf{Param Cnt} 
& \textbf{Time (min)} \\
\midrule
Full Fine-Tuning 
& 89.01 
& 100.0 
& 85,875,556 
& 39.46
 \\

LoRA (r=8)
& 89.93%
& 100.03 
& 1,327,104 
& 44.52 
 \\

DoRA (r=8)
& 89.97
& 101.08
& 1,410,048 
& 81.02
 \\

AdaLoRA ($12 \xrightarrow{}8$)
& 90.16 
& 101.29
& 1,991,520 
& 54.00  \\
\bottomrule
\end{tabular}
}
\end{table}

\bibliographystyle{IEEEtran}
\bibliography{references}
\end{document}